\documentclass[journal]{IEEEtran}
\def\@IEEEclspkgerror{\ClassError{IEEEtran}}
\usepackage{enumitem}
\usepackage{cite}
\usepackage{graphicx}
\usepackage{algorithm}
\usepackage{amsmath}
\usepackage{amssymb}
\usepackage{algpseudocode}
\usepackage{amsthm}

\usepackage{soul}
\usepackage[dvipsnames]{xcolor}
\usepackage{hyperref}
\usepackage[dvipsnames]{xcolor}

\usepackage{array}
\newcolumntype{P}[1]{>{\centering\arraybackslash}p{#1}}

\newcolumntype{M}[1]{>{\centering\arraybackslash}m{#1}}

\makeatletter
\newcounter{parenttheorem}

\makeatother

\def\BibTeX{{\rm B\kern-.05em{\sc i\kern-.025em b}\kern-.08em
    T\kern-.1667em\lower.7ex\hbox{E}\kern-.125emX}}
\usepackage{lineno}

\usepackage{float}
\floatstyle{plaintop}
\restylefloat{table}

\begin{document}

\title{{Deep Learning for Inertial Sensor Alignment}}
\author{Maxim Freydin, \IEEEmembership{Member, IEEE}, Nimrod Segol, Niv Sfaradi, Areej Eweida, and Barak Or, \IEEEmembership{ Member, IEEE}
    \thanks{Revision submitted: January 2024. Accepted: April 2024.}
\thanks{All authors are with ALMA Technologies Ltd, Haifa, 340000, Israel (e-mail: \{barak,maxim\}@almatechnologies.com).}}

\markboth{Deep Learning for Inertial Sensor Alignment / Preprint}%
{}
\maketitle

\begin{abstract}
Accurate alignment of a fixed mobile device equipped with inertial sensors inside a moving vehicle is important for navigation, activity recognition, and other applications. Accurate estimation of the device mounting angle is required to rotate the inertial measurement from the sensor frame to the moving platform frame to standardize measurements and improve the performance of the target task.  In this work, a data-driven approach using deep neural networks (DNNs) is proposed to learn the yaw mounting angle of a smartphone equipped with an inertial measurement unit (IMU) and strapped to a car. The proposed model uses only the accelerometer and gyroscope readings from an IMU as input and, in contrast to existing solutions, does not require global position inputs from global navigation satellite systems (GNSS). To train the model in a supervised manner, IMU data is collected for training and validation with the sensor mounted at a known yaw mounting angle, and a range of ground truth labels is generated by applying a random rotation in a bounded range to the measurements. The trained model is tested on data with real rotations showing similar performance as with synthetic rotations. The trained model is deployed on an Android device and evaluated in real-time to test the accuracy of the estimated yaw mounting angle. The model is shown to find the mounting angle at an accuracy of 8 degrees within 5 seconds, and 4 degrees within 27 seconds. An experiment is conducted to compare the proposed model with an existing off-the-shelf solution.

\end{abstract}


\section{Introduction}\label{sec:introduction}
\begin{figure}[ht]
\centering
{\includegraphics[width=0.47\textwidth]{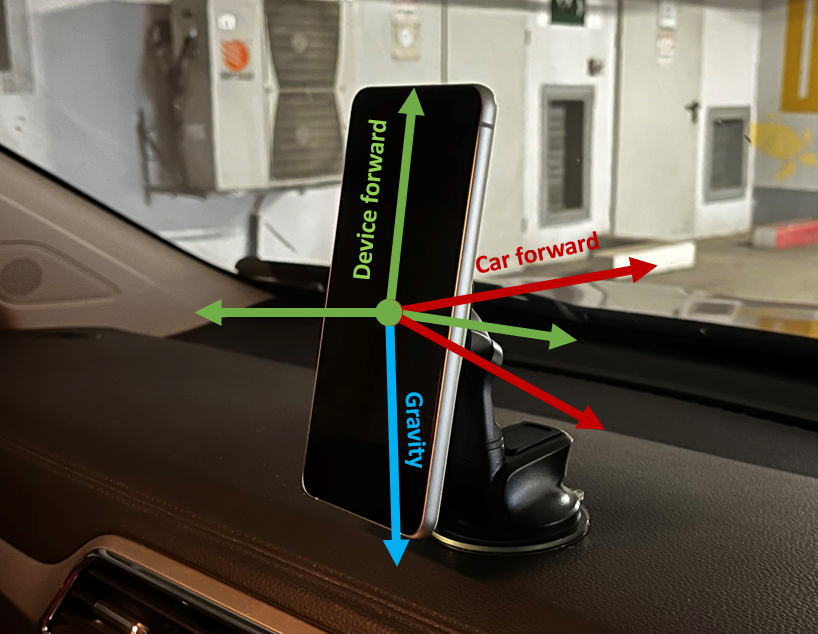}}
\caption{Smartphone device on a mount inside a car. Sensor data collected by the smartphone is measured in the sensor frame of reference (in green).}
\label{fig:Smartphone}
\end{figure}

\begin{figure}[ht]
\centering
{\includegraphics[width=0.35\textwidth]{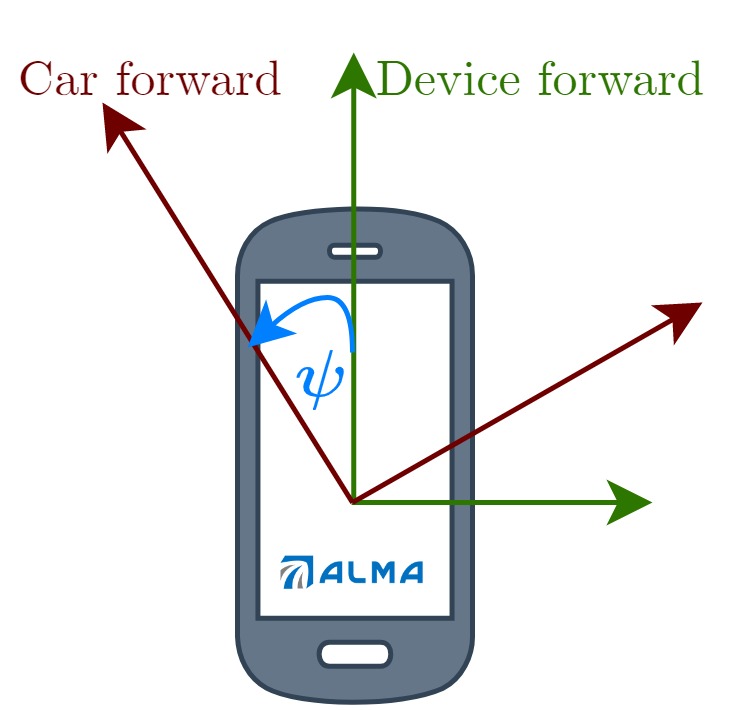}}
\caption{Top view schematic of the yaw mounting angle, $\psi$, after adjusting for roll and pitch angles with respect to the gravity vector.}
\label{fig:Smartphone_mounting_angle}
\end{figure}

\IEEEPARstart{I}{nertial} sensors are available on most mobile devices and are used in a wide range of applications including navigation \cite{dissanayake2001aiding,freydin2022learning}, {motion detection} \cite{seel2014imu, Jain2022}, activity recognition \cite{ashry2020}, {gait analysis} \cite{Gaud2023,semwal2022pattern},  and monitoring \cite{kashevnik2020cloud}, and other \cite{PENG2019247}. The inertial measurement unit, sometimes called a motion sensor, measures acceleration and angular velocity in the sensor frame at a sampling rate that depends on the hardware, typically between 100 and 500 Hz. Using IMU measurements has two notable properties: (1) the signals are noisy and (2) the acceleration and angular velocity vectors are obtained in the sensor frame. The noise is typically alleviated using classical signal processing methods and problem-specific modeling of the dynamics \cite{farrell2008aided}. However, transitioning between different frames of reference requires the relative orientation between frames which is unknown in most cases. In this work, a mobile device with an IMU is fixed to the dashboard of a car as shown in Figure \ref{fig:Smartphone} and the relative orientation between the IMU sensor and the car body is referred to as the mounting angle. More specifically, the problem is formulated such that only the yaw mounting angle is of interest, as shown in Figure \ref{fig:Smartphone_mounting_angle}.

Estimating and compensating for the yaw mounting (or alignment) angle of devices with a built-in IMU sensor is crucial {as it can drastically improve the performance of a wide range of tasks}. Indeed, the importance of detecting the IMU mounting angle was noted by several works \cite{hong2006experimental,niu2008improving,syed2008civilian}. In particular, \cite{syed2008civilian} conducted an experimental study on the effect of a misaligned sensor on the performance of INS/GNSS. Previous attempts to solve the mounting angle detection problem suffer from various ailments. Some methods use only acceleration data but require a calibration phase where the vehicle is horizontal \cite{mckown2017mounting,vinande2009mounting}. In \cite{pham2016detecting} data from motion sensors obtained while the vehicle is turning could not be used. Other approaches do not rely solely on IMU and require additional sensor inputs such as GNSS \cite{cheng2020accessibility,wu2010self, lu2022vehicle}. In \cite{chen2020estimate} the authors suggest a Kalman filter to estimate the mounting angle using IMU, GNSS, and odometer measurements. \cite{zhang2021mounting} presented a method using IMU and odometer to estimate the mounting angles. Recently, \cite{bassetti2022ml} attempted to learn the yaw angle using a synthetic data set generated similarly to this work. However, they used distinctly different machine learning algorithms (an ensemble of regressors) and obtained sub-par results compared to this work. In addition, their approach was not deployed or tested in real-time, which is a key goal of this work. Finding an IMUs mounting angle has also been studied outside the realm of position and navigation. The authors of \cite{chen2019imu} present an approach to estimate the mounting of pipeline inspection gauges. Their approach requires a calibration phase and utilizes accelerometer and gyroscope readings.  



In previous related work, DNNs were integrated into many navigation tasks. One of the first works in the field is the robust IMU double integration, the RIDI approach \cite{yan2018ridi}, in which neural networks were trained to regress linear velocities from inertial sensors to constrain the accelerometer readings. In \cite{herath2020ronin}, the device orientation, in addition to the accelerometer and gyroscopes readings, was used as input to a DNN architecture to regress the user velocity in 2D that was then integrated to obtain a position. For land vehicle navigation, a DNN model was presented to overcome the inaccurate dynamics and observation models of wheel odometry for localization, named RINS-W \cite{brossard2019learning}. In \cite{or2023learning}, recurrent neural networks were employed to learn the geometrical and kinematic features of the motion of a vehicle to regress the process noise covariance in a linear Kalman filter framework. In \cite{liu2021vehicle}, DNN-based multi-models combined with an EKF were proposed to deal with GNSS outages. In \cite{or2022hybrid} it was demonstrated that integrating DNN into classical signal processing algorithms can boost navigation performance. To the authors' best knowledge, this is the first work to suggest an end-to-end solution based on DNNs to find the yaw mounting angle of an IMU sensor.


In this work, a DNN model (more specifically, a CNN) with a smoothing algorithm for real-time deployment is developed which receives as input a window of IMU measurements and outputs the estimated sensor yaw mounting angle. A data-driven approach to solve the problem is proposed by collecting recordings of IMU from multiple drives with the IMU sensor strapped to a car at a known yaw mounting angle of zero degrees. To create a rich training dataset, the recorded IMU samples are rotated to simulate a wide range of mounting angles. While data is collected with an external IMU sensor at a prescribed angle, the trained model is deployed and tested on an Android mobile device to demonstrate generalization. 
The model is validated and tested on data with real and simulated rotations. {To summarize, the following contributions are made}
\begin{enumerate}
\item A DNN model is designed and trained to estimate the yaw mounting angle of a smartphone fixed to a car using only IMU measurements as input. The training dataset includes more than 52 hours of driving in 136 sessions.
\item A method to synthesize ground truth labels is presented and validated. This approach can substantially reduce the data collection effort in future studies.
\item An algorithm is formulated to process the DNN output and apply the trained model in real-time. 
\item The trained model is tested on a validation set consisting of more than 7 hours of driving in 18 sessions. The proposed method is shown to converge in up to 30 seconds to an accurate estimate for the yaw mounting angle with a 4 degrees error.
\item An experiment is conducted to test the trained model with the proposed algorithm in a real setting on an Android device. In addition, results are compared to an existing off-the-shelf solution which is based on the classical fusion of GNSS with IMU.
\item The trained model is shown to detect sensor rotation in the middle of the drive at the same speed and accuracy as for the initial angle. 
\end{enumerate}


The rest of the paper is organized as follows: 
Section \ref{sec:dataset} presents the data processing and dataset creation. Section \ref{sec:MountNet} presents the proposed MountNet model architecture and loss function. Section \ref{sec:realtime} presents an algorithm for smoothing and detecting changes in device mounting angle during a drive. Section \ref{sec:results} gives the results and Section \ref{sec:conclusions} presents the conclusions of this work.

\section{Dataset Creation} \label{sec:dataset}

Data-driven approaches require data collection for offline training and validation of models. In this section, data collection and processing are described.

\subsection{Raw data collection}
Acceleration and gyroscope data were collected with a WitMotion (WIT) BWT61CL IMU sensor connected to a Windows PC through a serial port connection. Data was recorded at the maximum sensor rate of $100$ Hz. A total of $60$ hours were collected for this task in  $154$ distinct driving sessions conducted by $4$ different drivers in $12$ different cities. Of these drives, $136$ drives totaling $52.8$ hours were used for training and the remainder for validation. The IMU sensor was placed in the same position and orientation on the car dashboard in every drive such that it aligned with the direction of the drive. That is, raw data was collected at a zero mounting angle with a small error due to human factor, i.e., the nominal yaw angle is $\psi=0$ degrees. 

\subsection{Data processing}
\label{sec:processing}
The collected data consists of acceleration $[m/s^{2}]$ and angular velocity $[rad/s]$ in three axes. Before passing the data to MountNet, we apply the following pre-processing tasks:
\begin{enumerate}
\item Apply a low pass filter with a cut-off frequency of $10$ Hz. This is justified because the dynamics of a typical car lie in this bandwidth \cite{rajamani2011vehicle}.
\item Down-sample the data to $20$ Hz. This step allows us to be agnostic to the recording device's frequency.
\item Rotate the measured acceleration and angular velocity vectors to the navigation frame where the third component is parallel to the gravity vector \cite{farrell2008aided}. This step limits the mounting angle estimation problem to be only a function of the yaw angle as defined in Figure \ref{fig:Smartphone_mounting_angle}. That is, the gravity vector is used to determine the roll and pitch angles, while the car's forward direction determines the relative yaw angle. {Roll and pitch estimation was implemented as described in} \cite{Candan2021}.
\item {After applying the roll and pitch rotations,} subtract the gravity vector from the third acceleration component.
\end{enumerate}

The above steps are applied offline when training the model and online for using it in real-time. The procedure was designed to simplify the signal and reduce noise and other unnecessary information that is passed to the DNN model. 

\subsection{Synthetic datasets creation}
\label{sec:data_generation}

As described above, data was collected with the sensor aligned with the car's forward direction. To make effective use of the existing data, and reduce human factor errors, the nominal ground truth is always zero degrees. Following the data processing process described in Section \ref{sec:processing}, the IMU measurements are independent of the roll and pitch mounting angles, and a synthetic yaw mounting angle can be prescribed. 

\begin{figure}[ht]
\centering
{\includegraphics[width=0.49\textwidth]{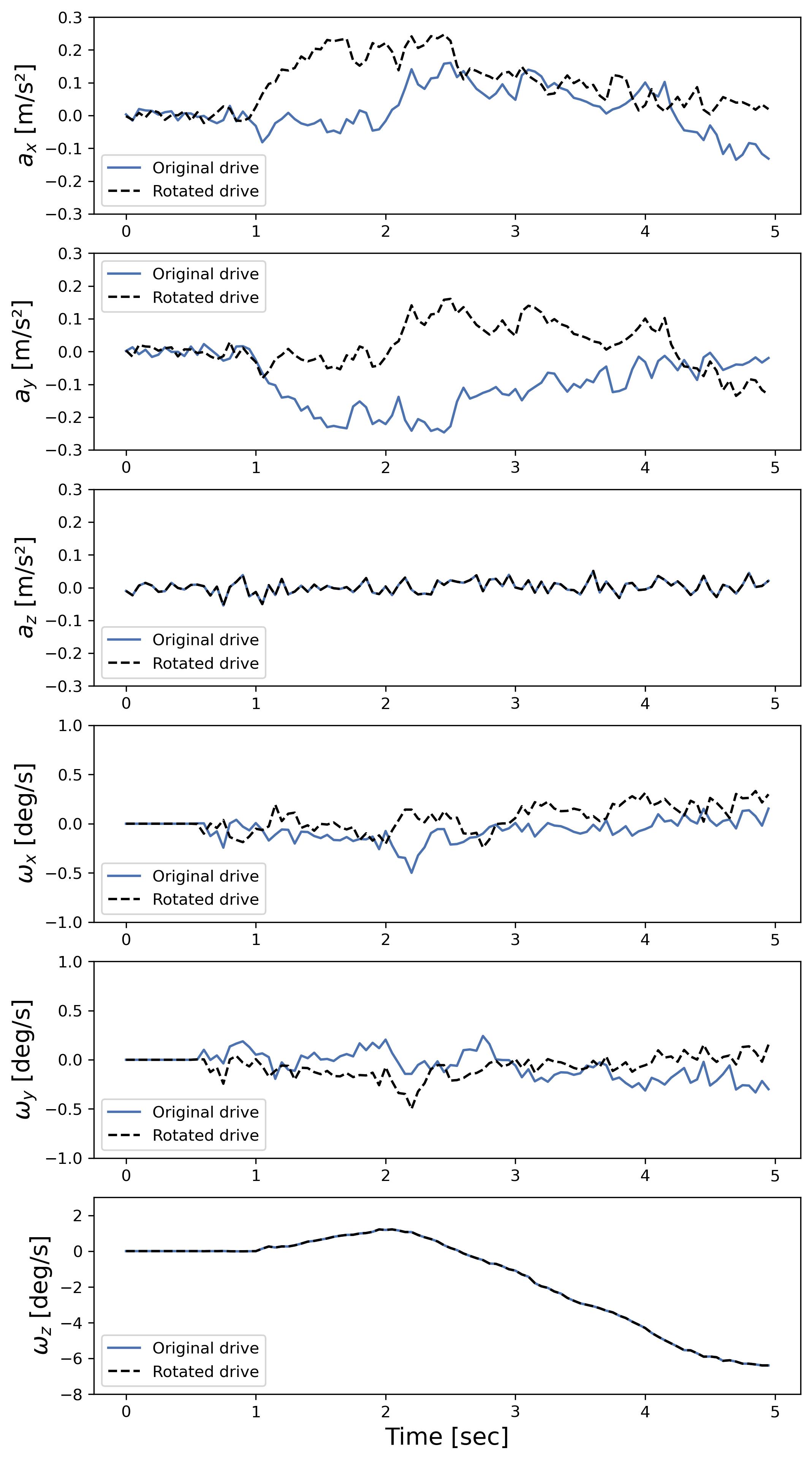}}
\caption{An example of a single sample in our dataset. We show both $x$ (in dotted black), the $5$ seconds of measurement from each of the IMU channels, and  $x_{\pi/2}$ (in blue) the same measurements when artificially rotated by an angle of $\pi/2$ about the $z$ axis as described in Section \ref{sec:data_generation}. We can see the effect the rotation has on each of the IMU channels (from top to bottom): acceleration in the x, y, and z axes and angular velocity in the x, y, and z axes.}
\label{fig:effect_of_rotation}
\end{figure}

To create the dataset, recorded drives were partitioned into windows of $5$ seconds with a $0.25$ seconds overlap. For each window, an angle $\psi$ is drawn from a uniform distribution in the range $[-\frac34\pi, \frac34\pi]$. This limits the problem to a typical range of mounting angles where the phone screen is facing the inside of the car. The processed IMU measurements window ${\bf x}\in \mathbb{R}^{100\times 6}$ is then rotated by the $\psi$ angle to synthesize a wide range of cases. The acceleration and gyroscope vectors are rotated about the gravity vector by applying a matrix product with
\begin{equation}
    {\bf \bar{R}}_{\psi} = \begin{pmatrix}
\cos\psi & -\sin\psi &0\\
\sin\psi & \cos\psi & 0\\
0 & 0 & 1
\end{pmatrix}  
\end{equation}
However, because the acceleration and gyroscope vectors are concatenated, we form the following block matrix,
\begin{equation}
{\bf R}_{\psi} = \begin{pmatrix}
{\bf \bar{R}}_{\psi} & {\bf 0}_{3\times 3}\\
{\bf 0}_{3\times 3} & {\bf \bar{R}}_{\psi} 
\end{pmatrix}.   
\end{equation}
The matrix product ${\bf x}_{\psi} = {\bf x}{\bf R}_{\psi}^{\top}$ rotates the nominal sample ${\bf x}$ by $\psi$ to produce the rotated sample ${\bf x}_{\psi}$. Figure \ref{fig:effect_of_rotation} shows an example sample rotation with $\psi = \frac{\pi}2$. Four of the six IMU channels ($z$ axis channels are the exception) are changed due to the simulated rotation.

\begin{figure}[ht]
\centering
{\includegraphics[width=0.25\textwidth]{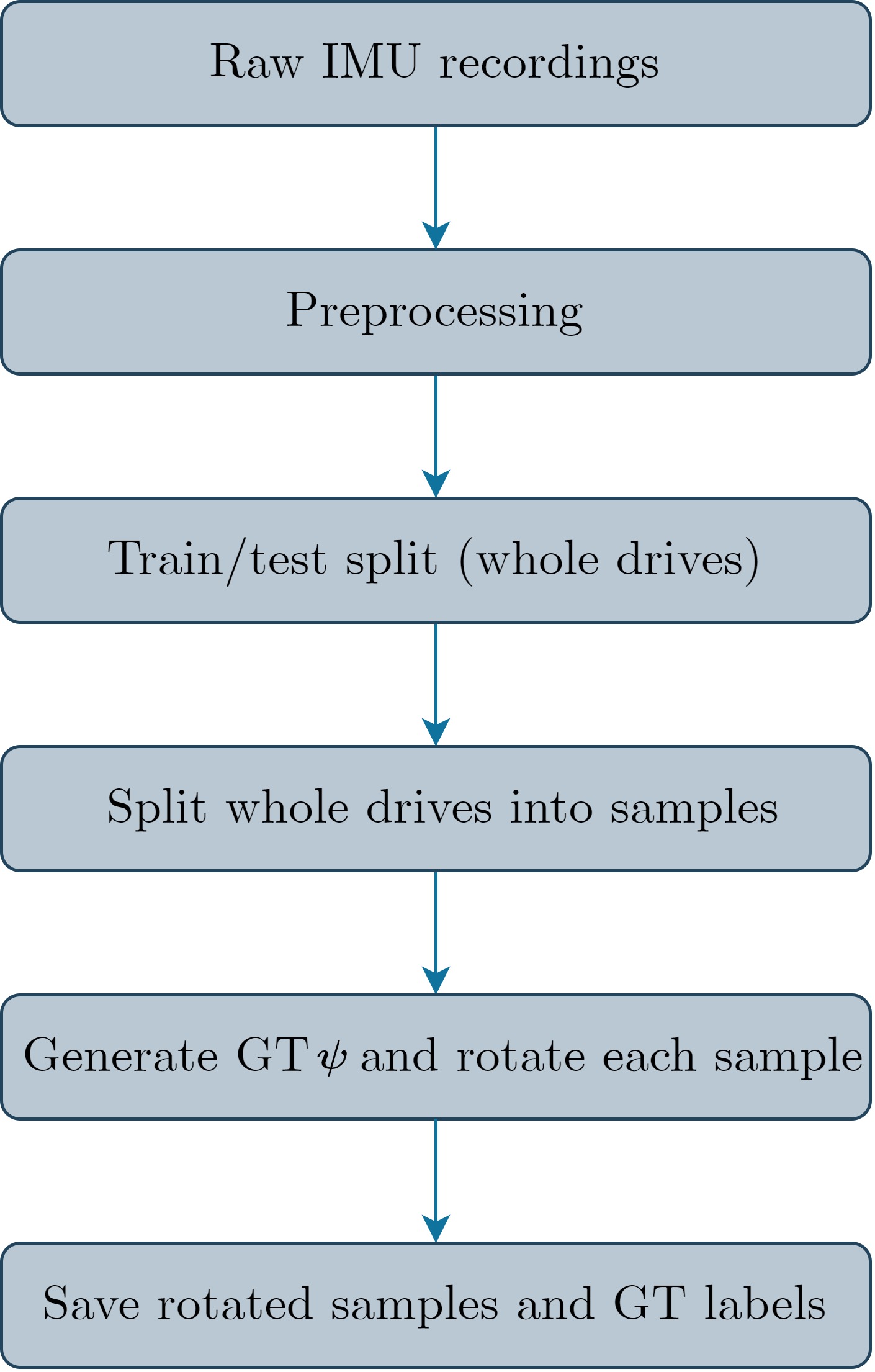}}
\caption{Data processing and datasets creation flowchart.}
\label{fig:pre_diagram}
\end{figure}

The dataset creation flowchart is shown in Figure \ref{fig:pre_diagram}. The dataset consists of pairs $(x_{\psi}, \psi)$ where $x\in\mathbb{R}^{100\times 6}$ is a window consisting of $5$ seconds of acceleration and gyroscope readings and $\psi\in \mathbb{R}$ is the simulated mounting angle. The training set consists of $672,858$ samples, and the validation set consists of $102,023$ samples, keeping a ratio of 85:15 in the train-validation split. 

\section{Model Architecture and Training}
\label{sec:MountNet}

The MountNet model is trained in a supervised manner with the dataset created as described above. In this section, the loss function, model architecture, and training regime are described.


\subsection{Loss function}
Usually, when dealing with regression tasks DNNs are trained with the mean squared error (MSE) loss. Namely, if the output of the DNN is $\tilde{\psi}(x, { \bf W})$ where $\bf W$ are the weights and $x$ is the processed IMU input, then the MSE is, 
\begin{equation}\label{eq:loss_mse}
{\bf MSE} \left( {\psi ,\tilde \psi \left(x,  {\bf{W}} \right)} \right) = \frac12{\left( {\psi  - \tilde \psi \left(x,  {\bf{W}} \right)} \right)^2}.
\end{equation}
For our purposes, the MSE loss presents a challenge since angles are periodic in $2\pi$. For example  ${\bf MSE}(0, \frac32 \pi) = \frac94\pi^2$, though the squared distance between the angles should be  $\frac14\pi^2$. To mitigate this issue, a loss function that is similar to the MSE but is agnostic to the periodicity of angles is considered. Specifically, the model is trained with a loss based on the function
\begin{equation}
\label{eq:loss}
\ell\left( \psi, \tilde{\psi}\left(x, \bf{W}\right)\right) =  1-\cos\left( {\psi  - \tilde \psi \left(x,  {\bf{W}} \right)} \right).
\end{equation}
During the training procedure, $L_2$ regularization with parameter $\lambda = 10^{-6}$ was applied so that the loss reads, 
 \begin{equation}
{\cal L} = \frac{1}{N} \left[\sum\limits_{i = 1}^N {\ell(\psi_i, \tilde{\psi}\left( x_i, \bf{W}\right)) }  \right]  + \lambda \left\| {\bf{W}} \right\|_2^2
\end{equation}
where N is the number of samples and $\ell$ is defined in Equation \ref{eq:loss}. Additionally, we trained the model using the standard MSE loss with L2 regularization. Results for the model trained with both losses are presented in Table \ref{table:1}. As can be seen, the performance of the entire algorithm is similar for both choices of the loss function. 

\subsection{MountNet Architecture }
A convolutional neural network architecture was used to solve the regression problem. This allowed the model to capture the temporal information in the IMU signal. the MountNet architecture layers are described here and presented in Figure \ref{fig:CNN}:

\begin{figure}[!ht]
\centering
{\includegraphics[width=0.47\textwidth]{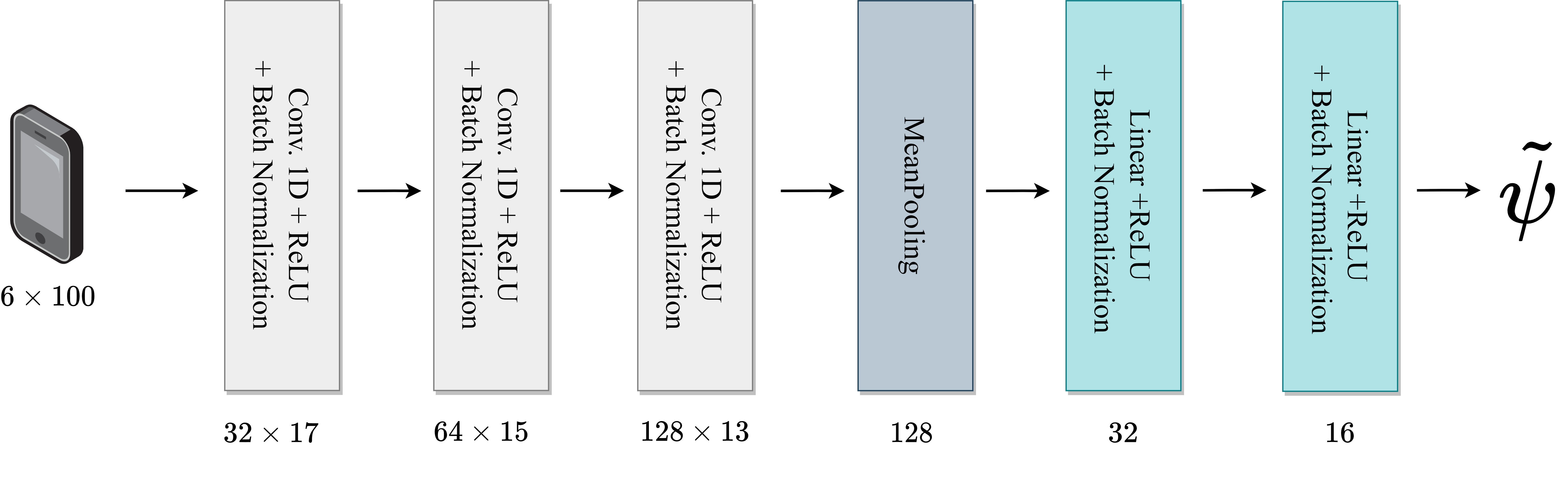}}
\caption{DNN architecture: three convolutional layers are followed by two dense layers with a connecting mean-pooling layer. The MountNet outputs the mounting angle of the smartphone device inside a car.}
\label{fig:CNN}
\end{figure}


\begin{enumerate}
\item {\bf Linear layer}: Applies a linear transformation to the input data from the previous layer. 
\item {\bf Conv1D layer}: A convolutional, one-dimensional layer creates a convolution kernel that is convolved with the layer input over a single dimension to produce a vector of outputs. In the MountNet model, the input layer is followed by a chain of three Conv1D layers. 
The dimensions of the convolutions layers  are $32$, $64$, and $128$ with kernel sizes $20$, $3$ and $3$ and stride $5$, $1$, and $1$, respectively.
\item {\bf Mean pooling}: Pooling layers reduce the dimension of feature vectors. In the mean pooling layer, the mean value for each input channel is calculated and flattened to be passed to the linear layers. 
\item {\bf ReLU}: Rectified linear unit is a nonlinear activation function with the output $ReLU (\alpha)= \max(\alpha, 0)$.
The ReLU functions were placed after every layer.
\item {\bf Batch normalization}:  Batch normalization reduces undesirable covariate shift. Batch normalization was added after every Conv1D layer (together with the ReLU layer). 
\end{enumerate}

\subsection{Model training}
The model consists of $39,905$ trainable parameters, which were initialized using Kaiming initialization \cite{He_2015_ICCV}. 
We trained the neural network for $1400$ epochs using the  ADAM optimizer \cite{kingma2014adam} with a learning rate of $0.001$, with $\beta$ coefficients of $\beta_1 = 0.9, \beta_2  = 0.999$, and batch size $128$.  We trained the model using a Tesla V100 Nvidia GPU. We trained for approximately 14 hours in total, 36 seconds on average per epoch. 

\section{Real-time Method}\label{sec:realtime}
The proposed approach is designed for application in real-time. In practice, the output of a trained MounNet mode described in Section \ref{sec:MountNet} presents some level of noise. {To address this issue, an algorithm is formulated to smooth a stream of the DNNs output. Given a trained MountNet model and a stream of IMU measurement batches collected at 2 Hz (a patch per 0.5 seconds), the algorithm returns the estimated sensor mounting angle at a rate of 2 Hz.}
 
\begin{figure}[!ht]
\centering
{\includegraphics[width=0.47\textwidth]{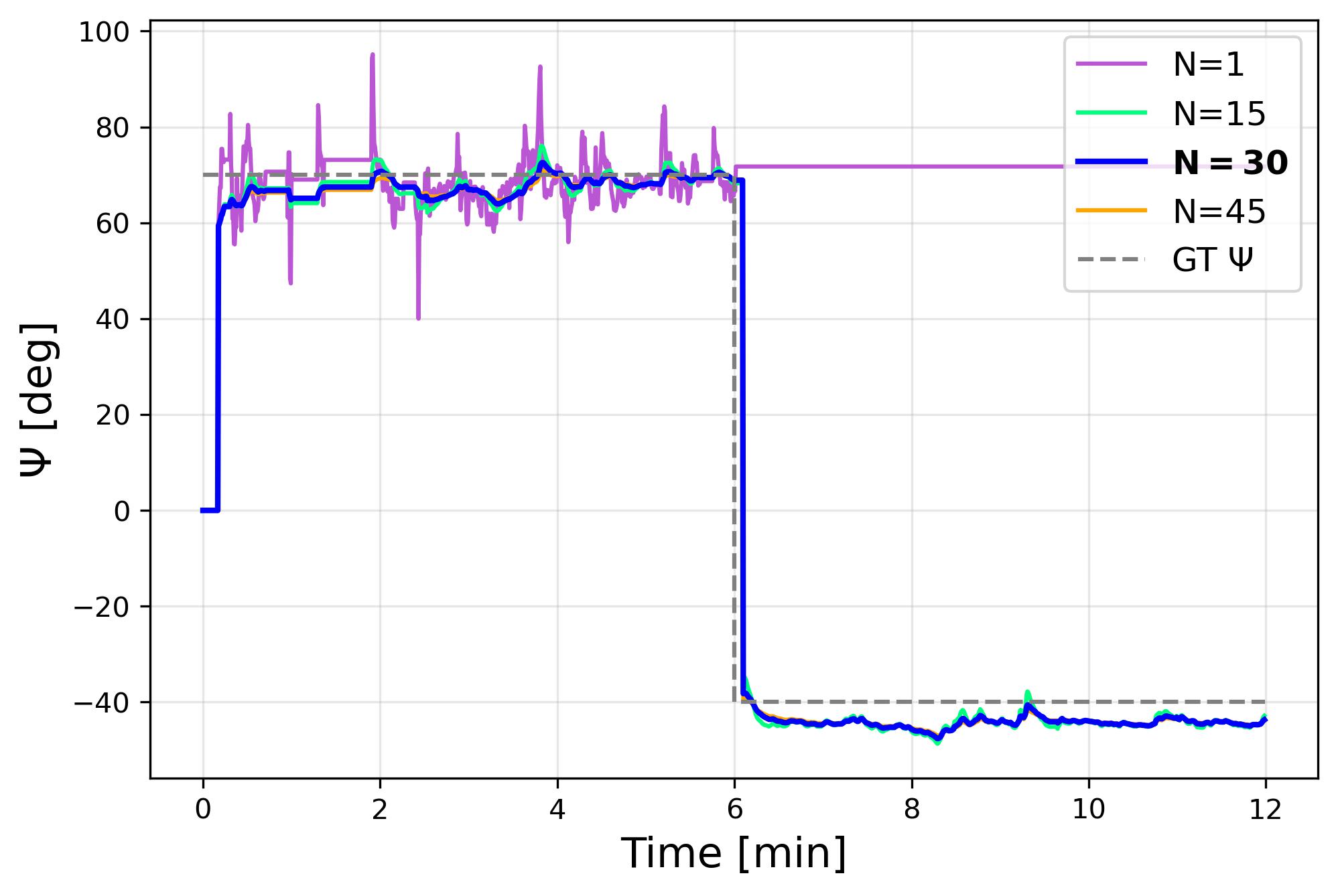}}
\caption{The result of Algorithm \ref{alg:realtime} for different values of smoothing window size $N$. For small smoothing window sizes, the output is noisy and less accurate. Without smoothing (N=1), the algorithm treats each output as an outlier and is unable to detect a change in the mounting angle mid-drive. With $N=45$, results are indistinguishable from $N=30$, despite the longer run time, supporting the choice of $N=30$. 
} 
\label{fig:smoothing_window_comparison}
\end{figure}

At deployment, a window of IMU measurements is passed every half a second through the processing procedure described in Section \ref{sec:processing}, and then the processed IMU data at time $t$, $x_t$, is fed to the MountNet model. After $5$ seconds of initial batch accumulation, the angle output is smoothed using a simple $\alpha-\beta$ filter \cite{brookner1998tracking}. The filter is formulated as,
\begin{equation}
\hat{\psi}_t = \frac {N-1}{N}\hat{\psi}_{t-1} + \frac 1N \tilde{\psi}\left(x_t, \bf{W} \right) \label{eq:smoother}
\end{equation}
where $N=\min(30, t)$ is { the chosen smoothing window size and $t$ is the current time,} $\hat{\psi}_{t-1}$ is the previous output of the algorithm, and  $\tilde{\psi}(x_t, \bf{W})$ is the output of the MountNet model at time $t$. 
We note that as $N$ increases the output of the algorithm becomes smoother but convergence time increases. We found that $N=\min(30, t)$ obtains a good trade-off between convergence time and smoothness, see Figure \ref{fig:smoothing_window_comparison} for a study of different smoothing window sizes. A detailed description of the algorithm appears in Algorithm \ref{alg:realtime}.


\begin{algorithm}
\caption{Real-time algorithm to estimate the mounting angle. Takes as input the MountNet model $\tilde{\psi}$ with trained weights ${\bf W}$ and processed IMU measurements. Outputs $\hat{\psi}_t$, the estimated mounting angle at time $t$. }
\label{alg:realtime}
\begin{algorithmic}[1]
\For{{$t=0[s], 0.5[s], 1[s], \ldots$}}  \, \# performed at $2$ Hz
\State $x_t \gets processed\ IMU\  data$  \, \# a batch of $0.5$ seconds
\If{$t < 5$}  \, \# initial padding
\State $\hat{\psi}_t \gets 0$
\Else
\State $X_t \gets concatenate(x_{t-5},...,x_t)$  \, \# Concatenate batches from the last 5 seconds
\State $\hat{\psi}_t \gets \tilde{\psi}\left(X_{t}, \bf{W}\right)$  \, \# Evaluate/inference step
\State $N \gets \min(t, 30)$
\State $\hat{\psi}_t \gets \frac 1N \hat{\psi}_t + \frac {N-1}{N}\hat{\psi}_{t-1};\quad Eq.  \eqref{eq:smoother}$
\EndIf

\EndFor
\end{algorithmic}
\end{algorithm}

\begin{figure}[ht]
\centering
{\includegraphics[width=0.47\textwidth]{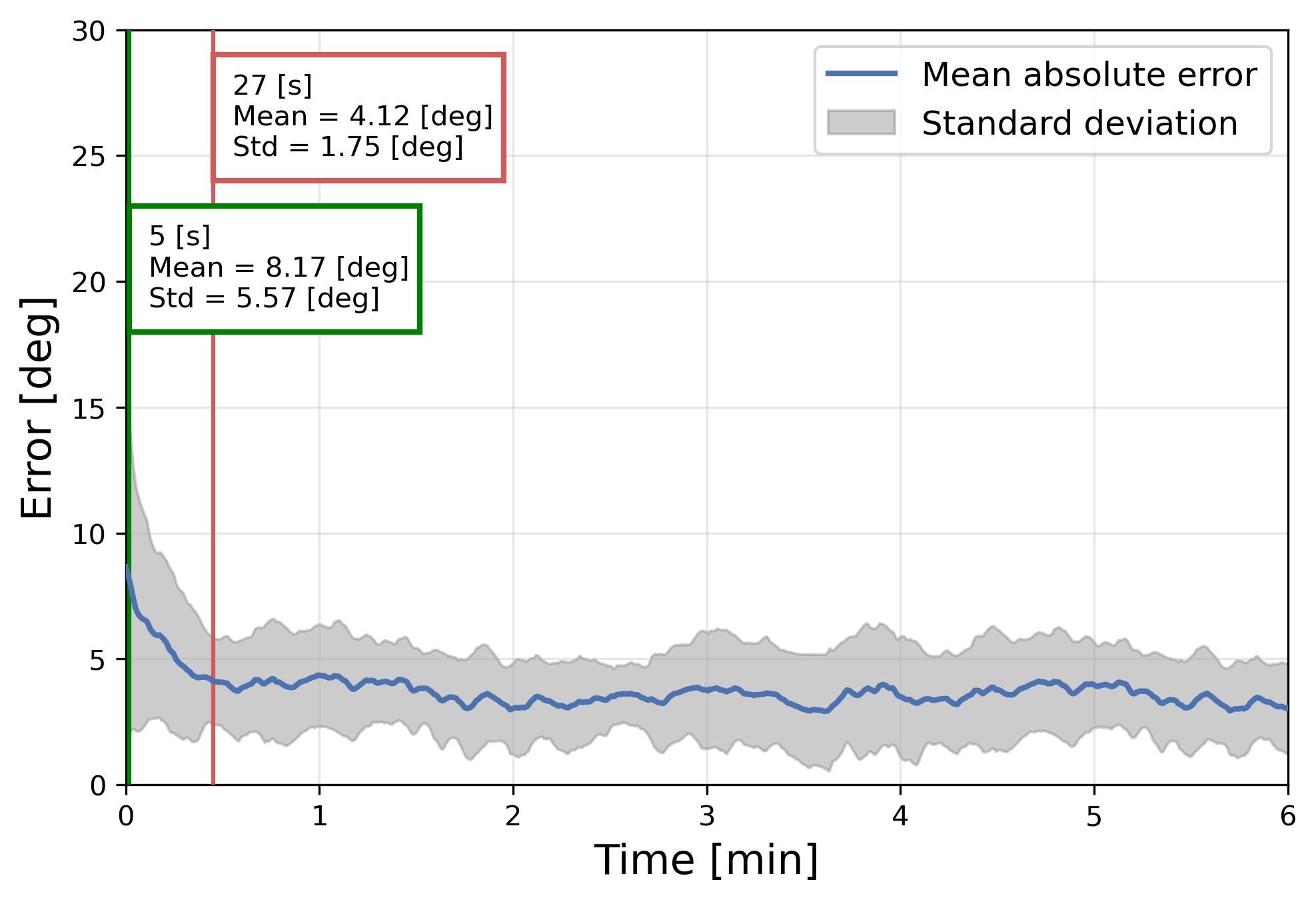}}
\caption{Estimated mounting angle mean absolute error and standard deviation band vs. time for the validation drives set. The proposed solution achieves a 4.12 degrees MAE after 27 seconds, and 3.5 degrees MAE after 90 seconds.}
\label{fig:mean_error_over_time}
\end{figure}

\section{Results}
\label{sec:results}
In this section MountNet performance is tested. Results are presented for: (1) a validation set collected with the external IMU and synthesized rotations, and (2) for a test set of recorded data using an Android Galaxy S21 device which was physically rotated (constant rotation and mid-drive rapid changes). In addition, an experiment was conducted where the performance of the proposed approach is compared to EVK-M8U \footnote{https://www.u-blox.com/en/product/evk-8evk-m8 (retrieved on May 2023)} (by u-blox), an off the shelf positioning product specializing in IMU/GNSS fusion. 

\begin{figure*}[ht]
\centering
{\includegraphics[width=0.98\textwidth]{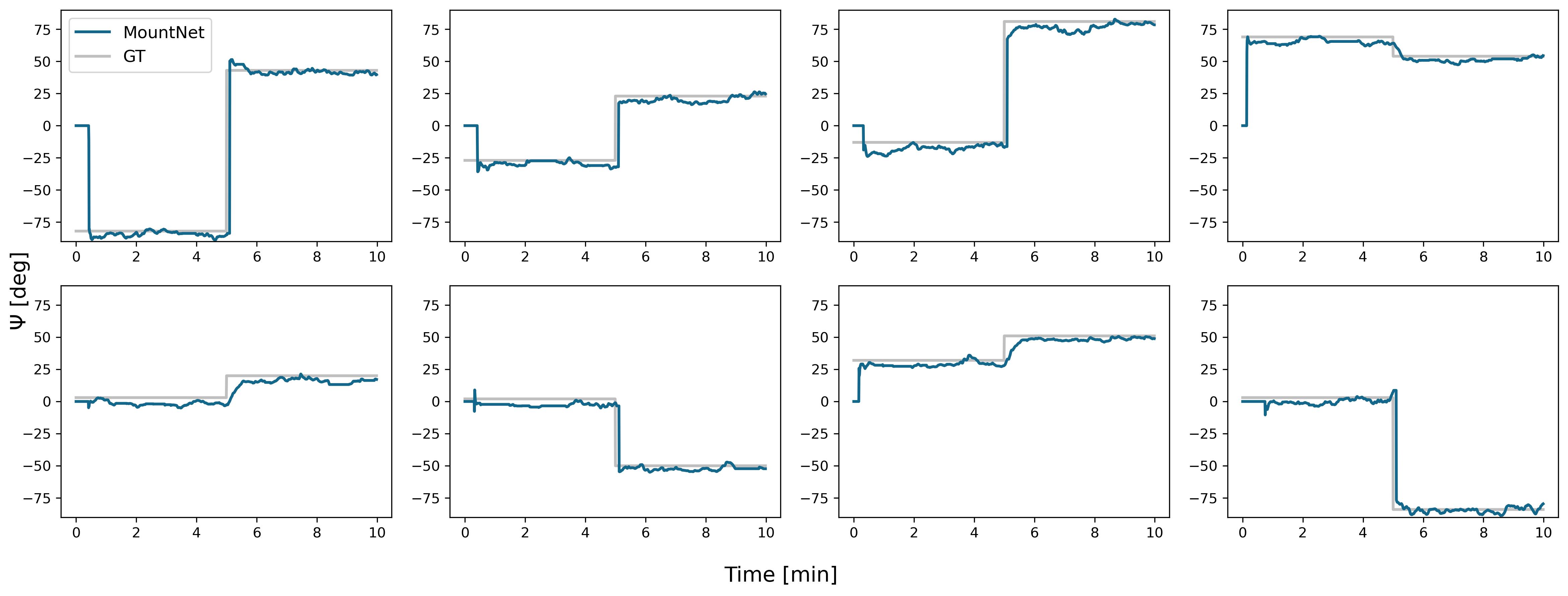}}
\caption{Estimated mounting angle vs. time with a synthetic angle change at $t = 5$ minutes.}
\label{fig:res_angle_change}
\end{figure*}


We follow the standard evaluation metrics for angle estimation: mean absolute error
(MAE) and root mean squared error (RMSE), as defined in Equation \ref{eq:metrics} for a set of size $\mathcal{M}$,

\begin{subequations}
\label{eq:metrics}
\begin{equation}
\label{eq:mae}
MAE = \frac{1}{{\cal M}}\sum\limits_{i = 1}^{\cal M} {\left| {{\psi _i} - {{\tilde \psi }_i}} \right|}
\end{equation}
\begin{equation}
\label{eq:rmse}
RMSE = \sqrt {\frac{1}{{\cal M}}\sum\limits_{i = 1}^{\cal M} {{{\left( {{\psi _i} - {{\tilde \psi }_i}} \right)}^2}} }.
\end{equation}
\end{subequations}

\subsection{Performance on the validation set}
\label{sec:validation_results}
Out of the total 60 hours of recorded data, 7 hours accounted for 18 driving sessions that were not used in the training process. IMU measurements from each of the 18 drives were processed as described in Section \ref{sec:processing} and a constant yaw rotation was applied to the full duration of each drive. The prescribed yaw angles were chosen randomly from a uniform distribution in the range of $[-\frac{\pi}{2}, \frac{\pi}{2}]$ radians. Performance metrics are summarized in Tabel \ref{table:1} for the raw MountNet output, the smoothed MountNet output using Algorithm \ref{alg:realtime}, and the smoothed output with the first minute of {driving} removed. The results show that the raw MountNet output is highly noisy as evidenced by the large difference between MAE and RMSE. It is also clear that smoothing improves substantially the accuracy of estimation. 


\begin{table*}[ht]
\centering
\begin{tabular}{| M{1.9cm} || M{1.7cm}| M{1.9cm} | M{1.9cm} |  | M{1.7cm}   | M{1.9cm} |   M{1.9cm} |   } 
 \hline
 \textbf{Metric} & \textbf{MountNet} & \textbf{MountNet w/smoothing} & \textbf{MountNet w/smoothing ($t > 1$ min)} & \textbf{MountNet - MSE}  & \textbf{MountNet w/smoothing - MSE} & \textbf{MountNet w/smoothing ($t > 1$ min)} - MSE \\ 
 \hline\hline
 MAE [deg] & 8.64 & 3.91 & 3.91&  12.51 &3.26 & 3.06\\ 
 \hline
 RMSE [deg] & 21.03 & 4.37 & 4.35 &24.07  &4.23 &3.85\\ 
 \hline
\end{tabular}
 \vspace{0.2cm}
\caption{Validation dataset performance metrics for MountNet raw output, smoothed output using Algorithm \ref{alg:realtime}, and smoothed output with the first minute of driving removed. On the left are results for the case when MountNet was trained using a periodic loss (Equation \ref{eq:loss}) and on the right with the MSE loss (Equation \ref{eq:loss_mse}). }
\label{table:1}
\end{table*}

The response convergence time of MountNet with Algorithm \ref{alg:realtime} was evaluated on the validation set. Figure \ref{fig:mean_error_over_time} shows the mean absolute error, $\left| {{\psi _i} - {{\tilde \psi }_i}} \right|$, of the combined algorithm with one standard deviation band versus time. The figure shows the performance of the combined algorithm during the first six minutes of the 18 validation drives. In the first 30 seconds, the output is relatively noisy however the standard deviation reduces rapidly after 15 seconds. The algorithm output error converges to approximately 4 degrees after 27 seconds and 3.5 degrees after 90 seconds of driving. {The run time of the combined algorithm was evaluated on an ASUS PC with Windows 10 and Intel Core i7 processor achieving a 0.5 millisecond per call (a single 5-second input batch).}




MountNet performance with Algorithm \ref{alg:realtime} was evaluated on the validation set with a second synthetic rotation during the drive. In this test, each validation drive is rotated by an additional randomly drawn yaw angle at $t = 5$ minutes.  Figure \ref{fig:res_angle_change} shows the output of the combined algorithm for 8 different drives. In all examples, the output converges near the true angle within 30 seconds of driving. This is true for the initial yaw angle at the drive start and after the angle change. It was observed that convergence after an angle change is reached within 5 seconds for changes greater than $\frac{\pi}{6}$. For smaller angle changes, convergence takes up to 30 seconds and the estimated $\psi$ changes gradually.

\begin{figure}[h!]
\centering
{\includegraphics[width=0.47\textwidth]{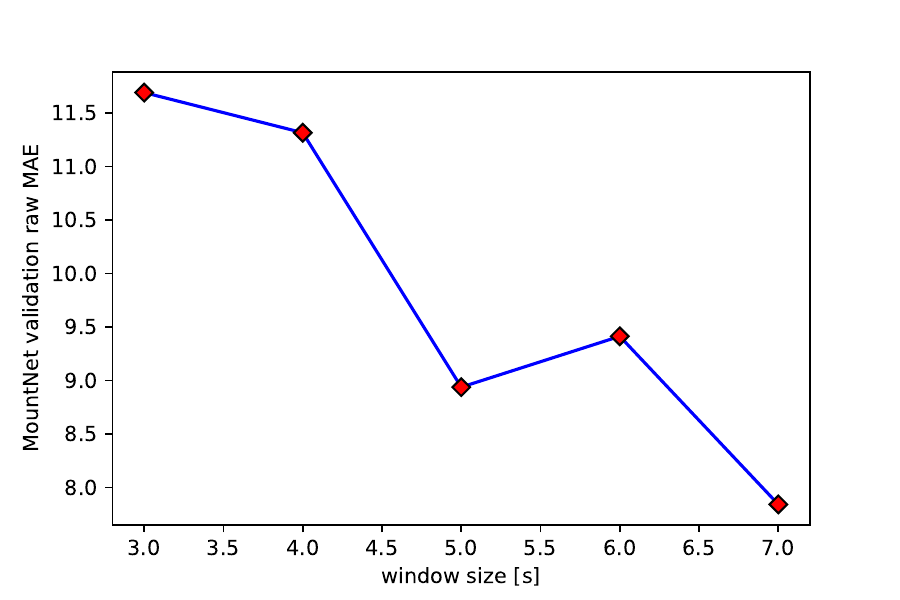}}
\caption{Validation MAE for MountNet trained with different input window size ranging from 3 to 7 seconds.}
\label{fig:window}
\end{figure}

The effect of input window size on model performance was investigated. As can be seen in Figure \ref{fig:window}, for increased window size, MountNet achieves better accuracy of the estimated angle on the validation set. A window size of 5 seconds was chosen by the authors to strike a balance between the improved performance of larger inputs and the computational cost and computational time of larger window sizes.

\subsection{Experiment results on a smartphone device}
\label{sec:phone_experiments}
\begin{figure}[h!]
\centering
{\includegraphics[width=0.47\textwidth]{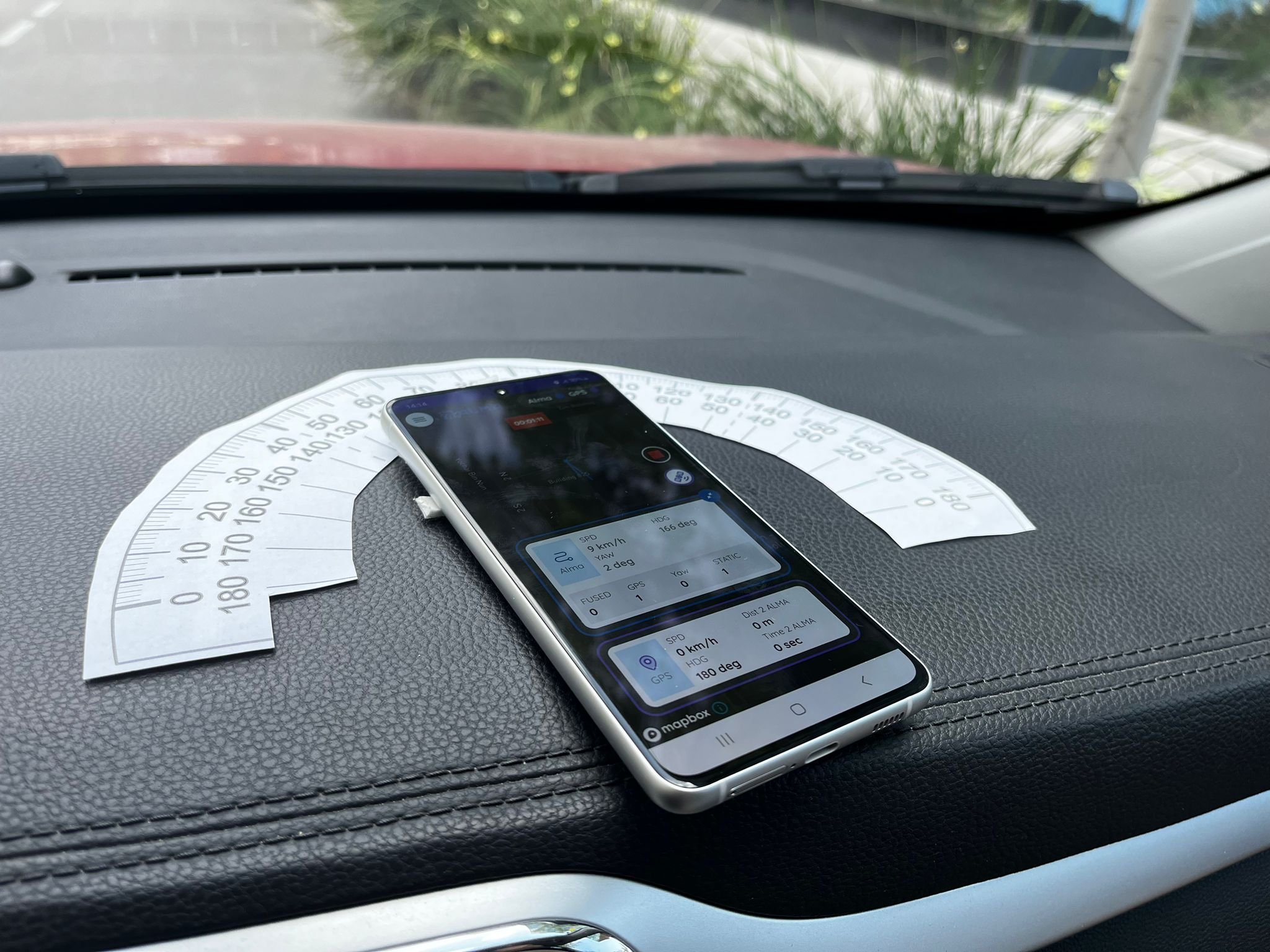}}
\caption{The Android Galaxy S21 device and protractor used to measure the ground truth mounting angle.}
\label{fig:goniometer}
\end{figure}

\begin{table}[ht!]
\centering
\begin{tabular}{| M{1.9cm} || M{1.7cm}| M{1.7cm} | M{1.7cm} |    } 
 \hline
 \textbf{Metric} &  10 seconds& 20 seconds &  30 seconds \\ 
 \hline\hline
 MAE [deg] &11.94 & 6.38 & 5.87 \\ 
 \hline
 RMSE [deg] & 14.66 & 8.52 & 6.44 \\ 
 \hline

\end{tabular}
 \vspace{0.2cm}
\caption{Smartphone mounting angle error after N seconds since start of driving.}
\label{table:2}
\end{table}

Experiments were conducted with a Galaxy S21 Android device {that has a built-in IMU sensor operating at a sampling rate of 500 Hz.} The trained model with the smoothing algorithm were deployed on a testing app with data logging and recording capabilities. The computations were performed in real-time and on-device. {Data processing and MountNet evaluation were performed at the rate of 2 Hz and each cycle took less than 15 milliseconds suggesting that higher rates are feasible.} In each drive the phone was physically rotated and the true angle was measured using a protractor as shown in Figure \ref{fig:goniometer}. 

Two experiments were conducted in this setting.  First, the smoothed output of MounNet was compared to the ground truth after driving time of $10$, $20$ and $30$ seconds for $9$ distinct drives at different device angles. The results are summarized in Table \ref{table:2}. It is noted that the model produces good results for a smartphone device physically rotated, even though it was trained on data from an external sensor glued to the dashboard with synthetic rotations for labels. The results in this experiment are comparable to those presented for the validation set.

\begin{figure}[h!]
\centering
{\includegraphics[width=0.5\textwidth]{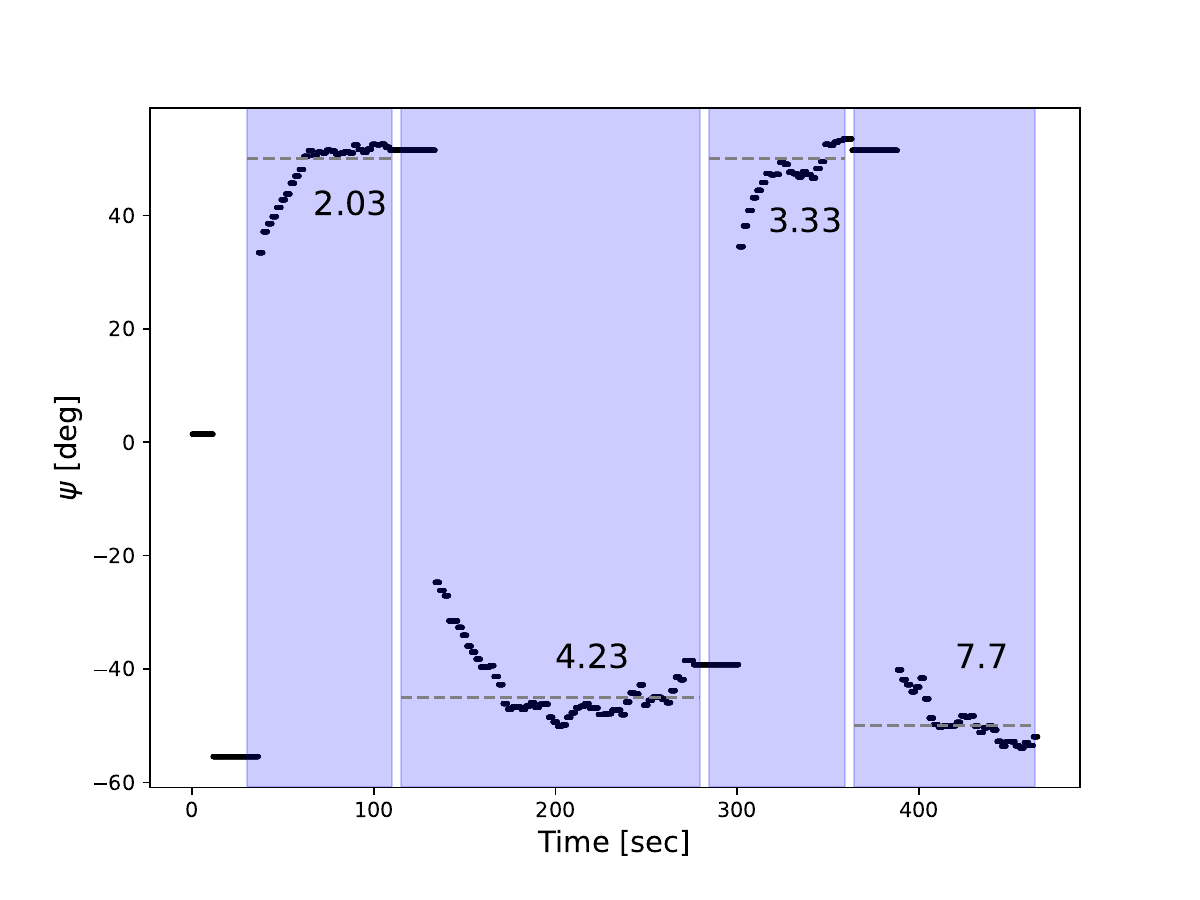}}
\caption{Smoothed MountNet response with mid-drive rotations of the mobile device. The dotted line is  the ground truth label. The parts in white represent a time when the smartphone was picked up. The average error between the estimation and ground truth after 20 seconds of driving at a new rotation angle is mentioned for each segment.}
\label{fig:phnemiddrive}
\end{figure}

The second experiment measures MountNet's ability to adjust to a change in the mounting angle mid-drive. A drive was recorded with the smartphone device mounted on the dashboard of a car and rotated several times without stopping the vehicle. As can be seen in Figure \ref{fig:phnemiddrive}, MountNet successfully captures the rotation and provides a continuous estimation of the device's mounting angle.  

\subsection{Experiment results with comparison to an existing solution}
\label{sec:field_experiments}
\begin{figure}[h!]
\centering
{\includegraphics[width=0.47\textwidth]{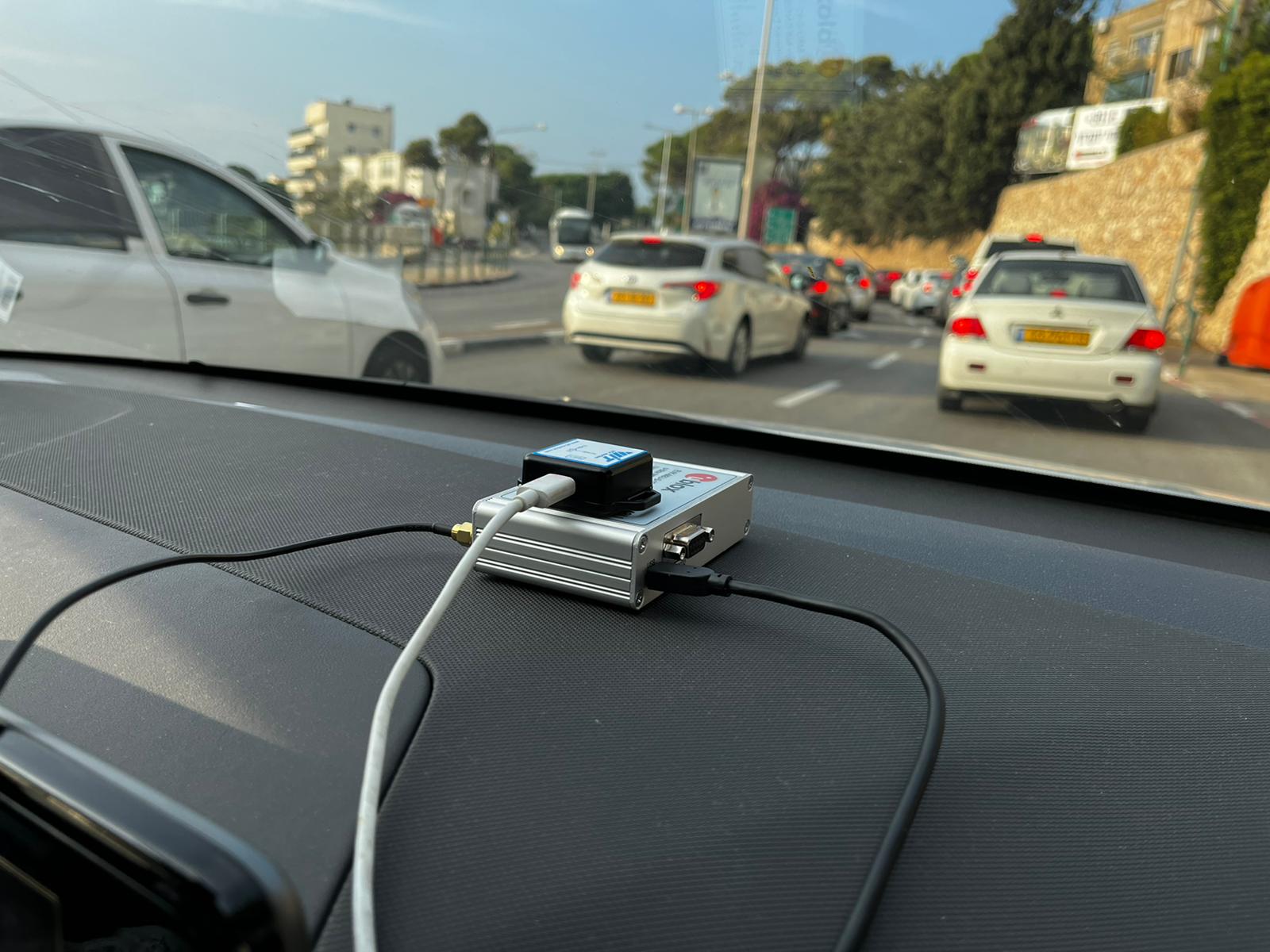}}
\caption{Experiment configuration for comparison with existing methods. The u-blox EVK-M8U evaluation device is fixed to the dashboard at near zero roll, pitch, and roll angles. The WIT IMU is fixed to and aligned with the u-blox device.}
\label{fig:ublox_imu_image}
\end{figure}

In this section, an experiment is conducted where a real rotation is applied to the external IMU sensor, and an EVK-M8U (by u-blox) device is used for comparison with an existing solution. EVK-M8U is an evaluation kit that fuses IMU and GNSS for dead reckoning applications in cars. Similarly to a mobile phone device with an IMU, the EVK-M8U is fixed to the car to sense the motion of the car. During its first minutes of operation, in the initialization phase, the u-blox device requires a high-quality GNSS signal to estimate internal parameters which include the yaw mounting angle. Figure \ref{fig:ublox_imu_image} shows the configuration of both devices where the external IMU (WitMotion) is fixed to and aligned with the EVK-M8U.

\begin{figure}[h!]
\centering
{\includegraphics[width=0.47\textwidth]{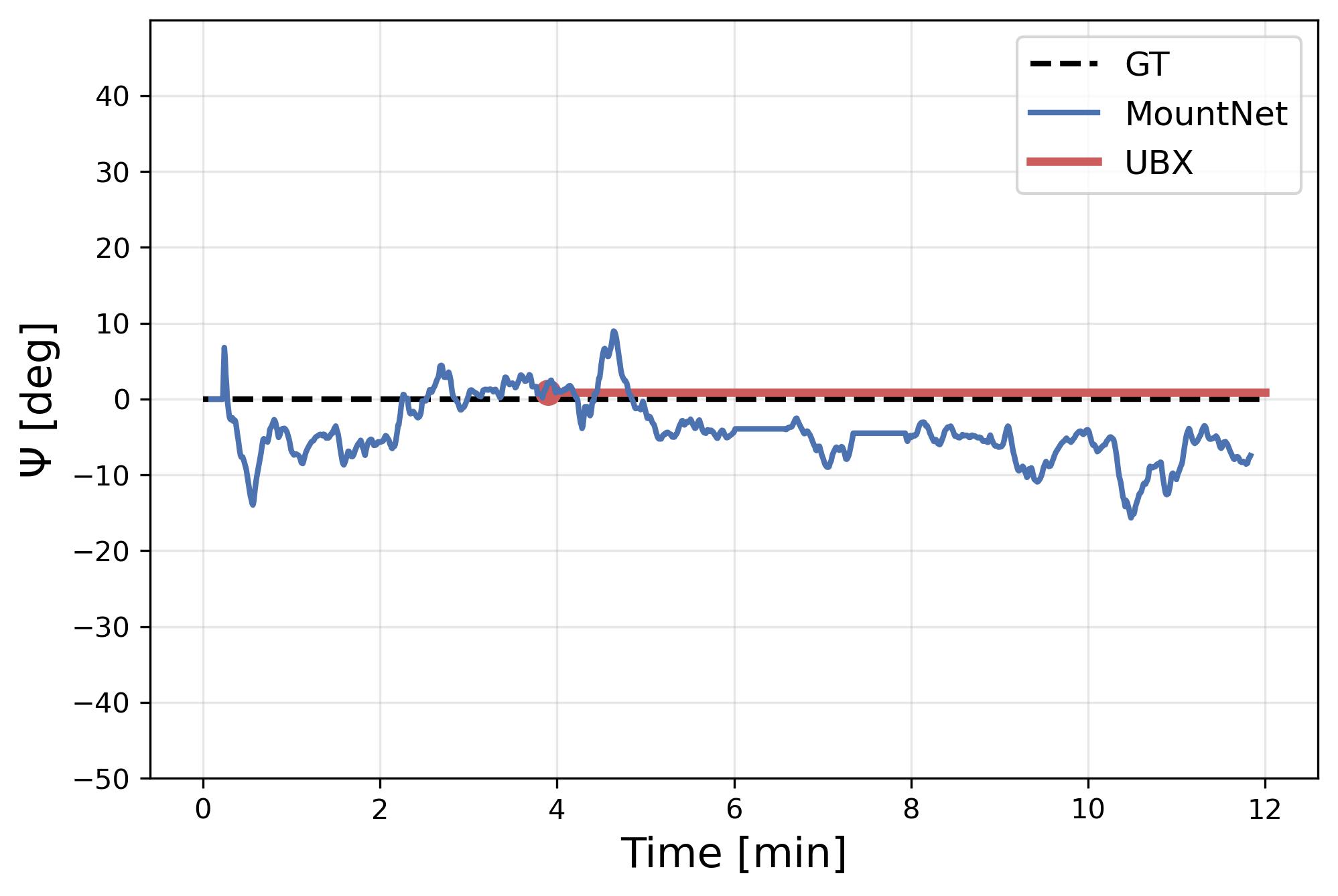}}
\caption{Estimated mount angle vs. time using MountNet and u-blox. MountNet MAE is 4.4 degrees.}
\label{fig:ublox_fixed_angle}
\end{figure}

In the first test drive, both sensors were aligned with the car's forward direction as shown in Figure \ref{fig:ublox_imu_image} (yaw mounting angle at zero degrees). MountNet was applied with Algorithm \ref{alg:realtime} and EVK-M8U output was monitored. Figure \ref{fig:ublox_fixed_angle} shows the output of both methods and the ground truth. The first output from EVK-M8U was received 4 minutes into the test with an estimated yaw mounting angle of $\psi={0.84}$ degrees. In the first 30 seconds of the drive, our approach resulted in an estimation error of 10 degrees. Over the whole drive, MountNet achieved an accuracy of 4.4 degrees MAE.

\begin{figure}[h!]
\centering
{\includegraphics[width=0.47\textwidth]{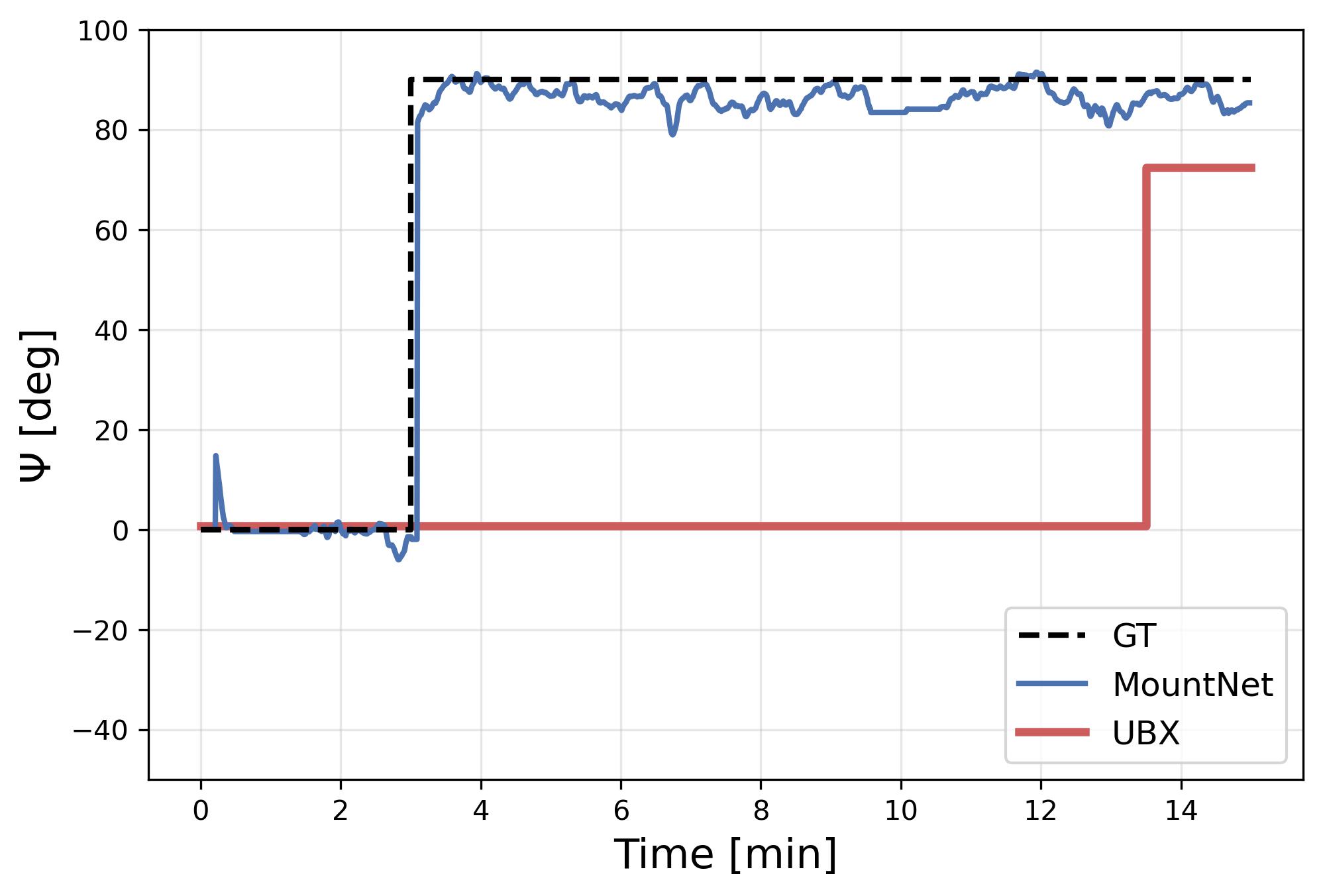}}
\caption{Estimated mount angle vs. time using MountNet and u-blox with angle change mid-drive. MountNet MAE is 0.8 degrees for the initial angle and 3.8 after the change. After the angle change, MountNet converges within 5 seconds.}
\label{fig:ublox_changing_angle}
\end{figure}

By the end of the first test drive, the EVK-M8U fully finished initialization. The second test drive started with a 3 minutes drive at zero yaw mounting angle followed by a 90 degrees counter-clockwise rotation at $t = 3$ minutes. Figure \ref{fig:ublox_changing_angle} shows the output from our method and the u-blox device during the second drive. In the first three minutes, both methods agree well, however, at the moment of rotation, the MountNet responds after 5 seconds to a highly accurate estimation with an MAE of 3.8 degrees while the EVK-M8U takes more than 10 minutes to respond. The final estimation of EVK-M8U converged to 70 degrees with an error of about 20 degrees. 



\section{Conclusions} \label{sec:conclusions}
{A data-driven approach was presented for estimating the yaw mounting angle of a fixed IMU sensor with respect to the body axis of a car. Existing solutions for mounting angle estimation are based on classical fusion algorithms of IMU and global position updates from an additional sensor, usually based on GNSS, and require several minutes to obtain a reliable estimation. The proposed solution requires only IMU as input and provides accurate results within seconds thus enabling a range of real-time applications on mobile devices including navigation, motion tracking, and activity recognition. This work focused on the formulation of the supervised training problem, data processing, deployment for real-time use, and conducting experiments and tests. The following are the key conclusions of this work:}

\begin{enumerate}
    \item {It was shown that the training data can be manipulated to generate a rich set of examples while reducing significantly the data collection effort. The trained model was shown to perform well on a synthetic validation set and in a real setting.}
    \item {Generalization was demonstrated from a low-cost external IMU sensor to a mobile smartphone device with a built-in IMU. The native sampling rates of the two sensors were different as well as the acquisition interface (serial port vs. Android OS).}
    \item {The proposed solution was found to be computationally light, relatively easy to deploy, fast to converge, and sufficiently accurate for a wide range of applications. Approximately 5 seconds of driving is necessary to provide an estimation within 4 degrees of the real angle which is an order of magnitude less time than classical techniques.}
    \item {A comparison with an existing solution in the form of an off-the-shelf product shows comparable results in accuracy but without dependency on a GNSS sensor and requiring significantly less time of driving.}
\end{enumerate}

{Future work will focus on extending the present approach to learn the roll, pitch, and yaw mounting angles simultaneously to provide a complete solution for sensor-to-body frame alignment. This work demonstrated the feasibility and strengths of data-driven approaches for sensor alignment and future work will emphasize DNN architecture optimization. It is expected that deeper architectures may be needed to solve the complete alignment which includes all three rotations. In addition, future work will focus on models which are capable of evaluating the quality of their estimation. For example, when a vehicle is stationary, the IMU measurements do not hold information about the mounting angles and thus the output of an aware model will include an indication that the current output is not reliable and should not be used.}


\bibliographystyle{IEEEtran}
\bibliography{IEEEfull}

\begin{IEEEbiography}[{\includegraphics[width=1in,height=1.25in,clip,keepaspectratio]{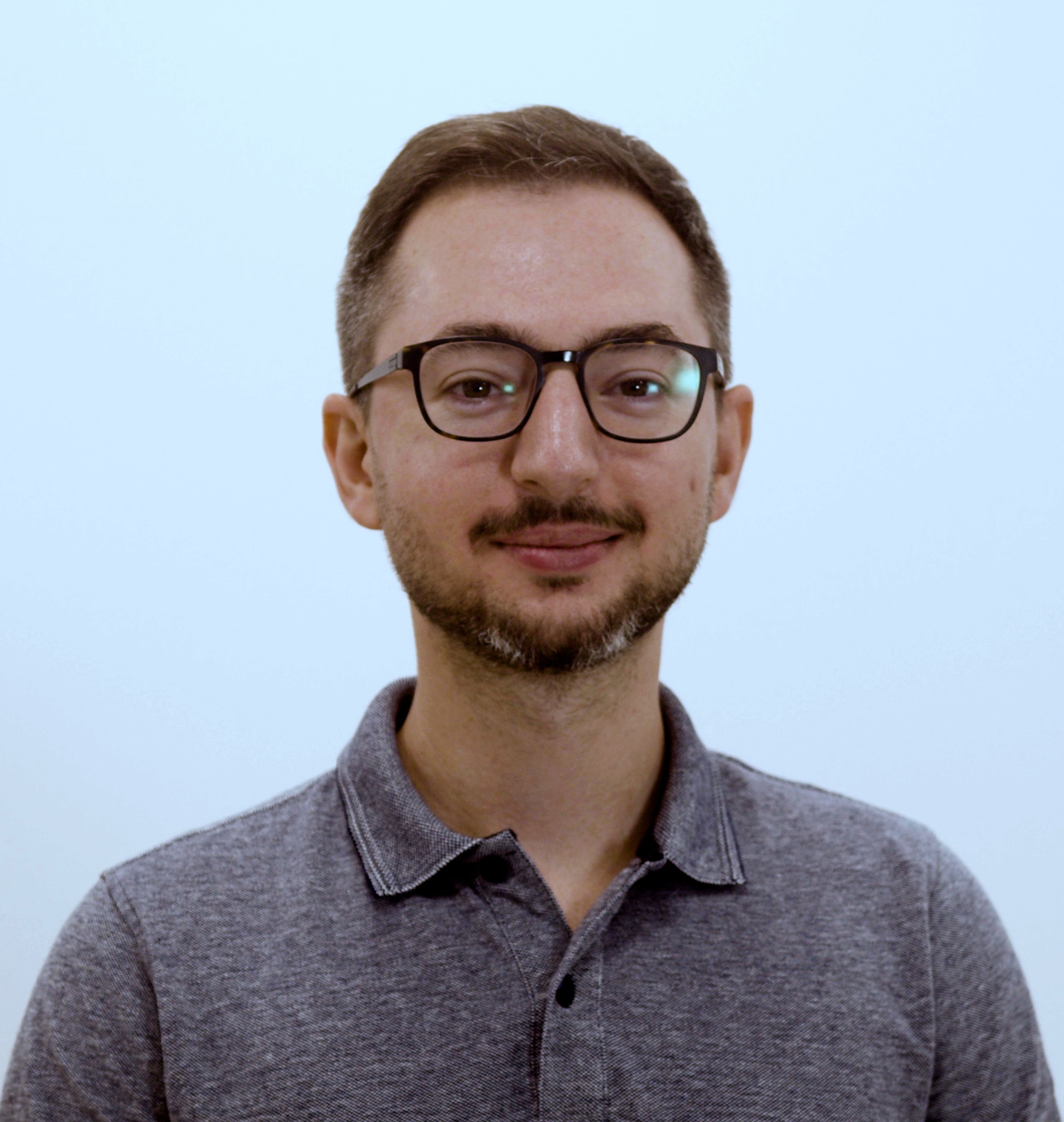}}]{Maxim Freydin} (Member, IEEE) received his B.Sc. (2017, Summa Cum Laude) and Ph.D. (2021) degrees in Aerospace Engineering from the Technion – Israel Institute of Technology. He is the VP of R\&D at ALMA Technologies Ltd. and his research interests include navigation, signal processing, deep learning, and computational fluid-structure interaction.
\end{IEEEbiography}

\begin{IEEEbiography}[{\includegraphics[width=1in,height=1.25in,clip,keepaspectratio]{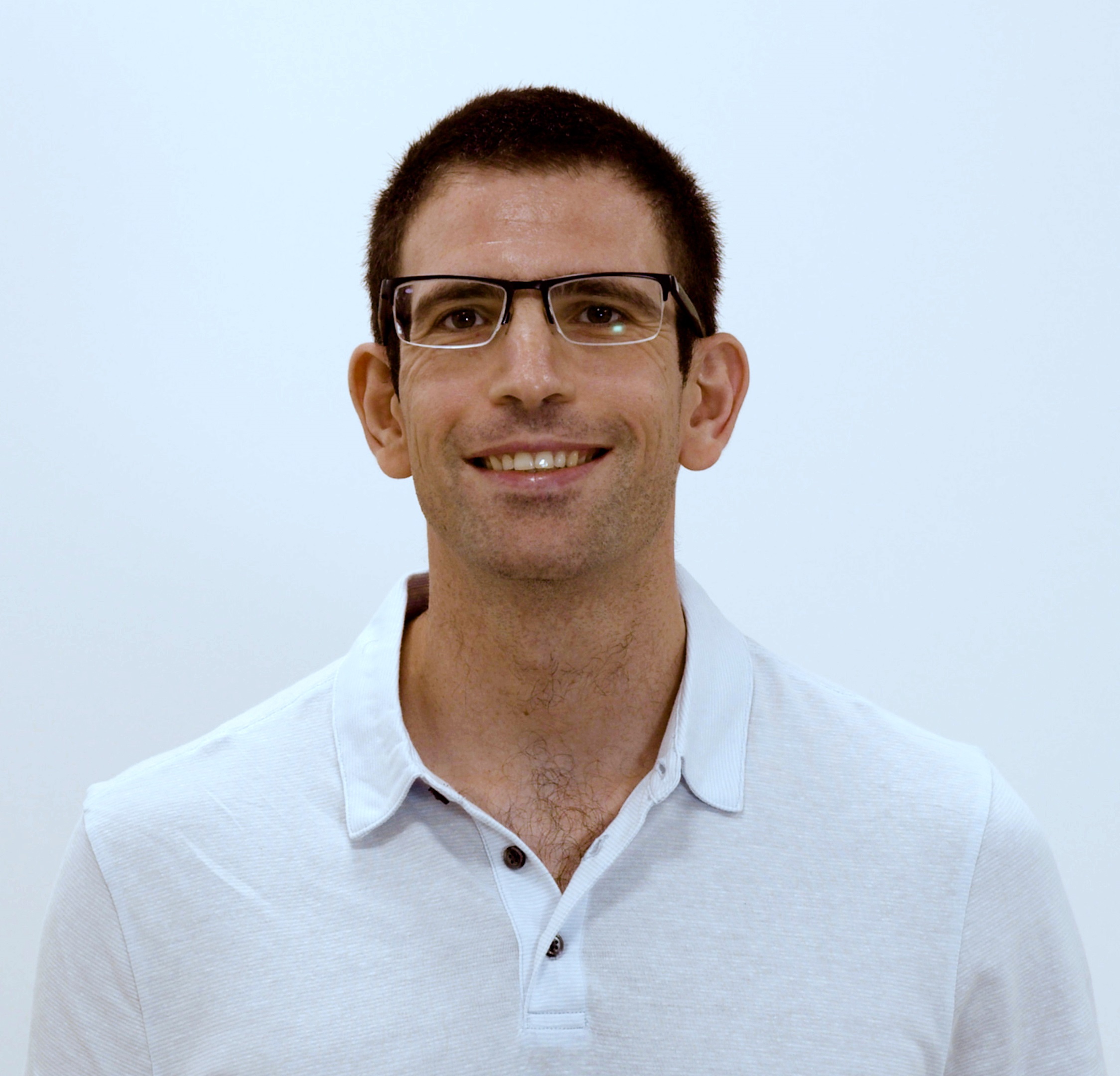}}]{Nimrod Segol}  received his B.Sc. (2015,  Cum Laude) and M.Sc. (2018) degrees in Mathematics from the Technion – Israel Institute of Technology. He is an algorithm engineer at ALMA Technologies Ltd. and his research interests include deep learning, machine learning, and statistics.
\end{IEEEbiography}

\begin{IEEEbiography}[{\includegraphics[width=1in,height=1.25in,clip,keepaspectratio]{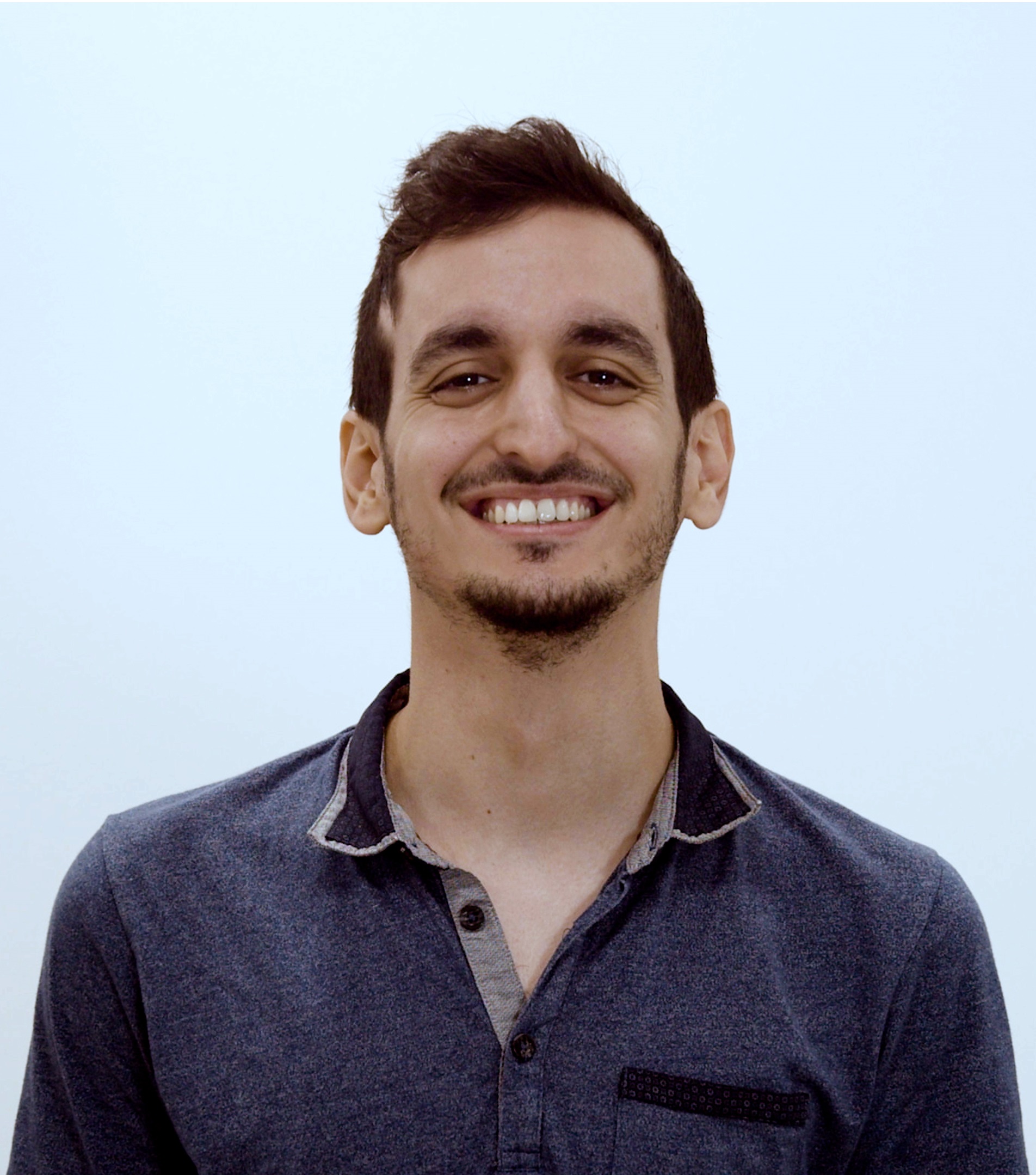}}]{Niv Sfaradi} received his Data Science and Engineering B.Sc. from the Technion - Israel Institute of Technology. He is an algorithm engineer at ALMA Technologies Ltd. and his research interests include deep learning, machine learning, and NLP. 
\end{IEEEbiography}

\begin{IEEEbiography}[{\includegraphics[width=1in,height=1.25in,clip,keepaspectratio]{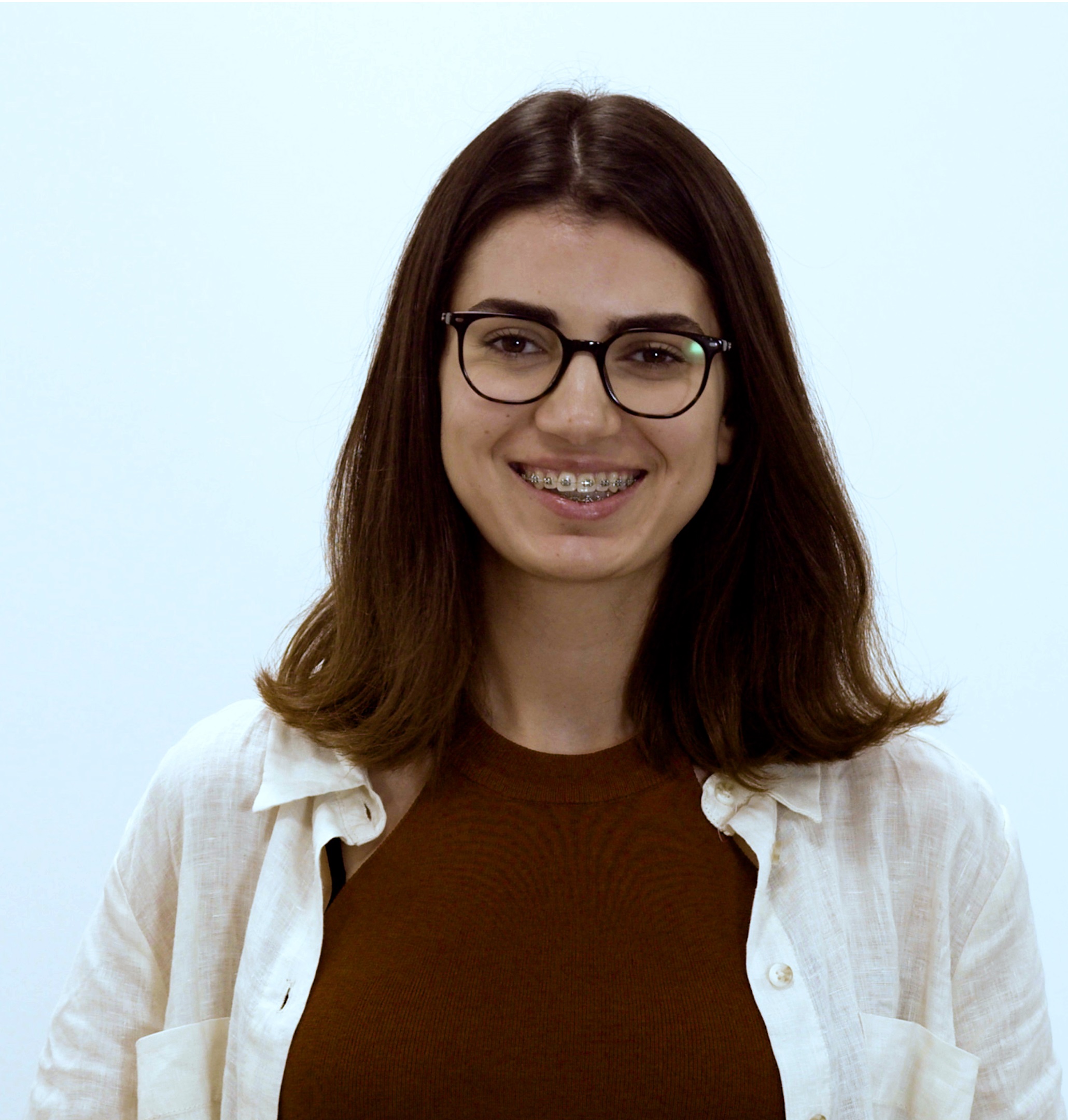}}]{Areej Eweida} received her B.Sc. (2023) degree in Aerospace Engineering from the Technion – Israel Institute of Technology. She is an algorithm engineer at ALMA Technologies Ltd. Her research interests include estimation theory, signal processing, and deep learning.
\end{IEEEbiography}

\begin{IEEEbiography}[{\includegraphics[width=1in,height=1.25in,clip,keepaspectratio]{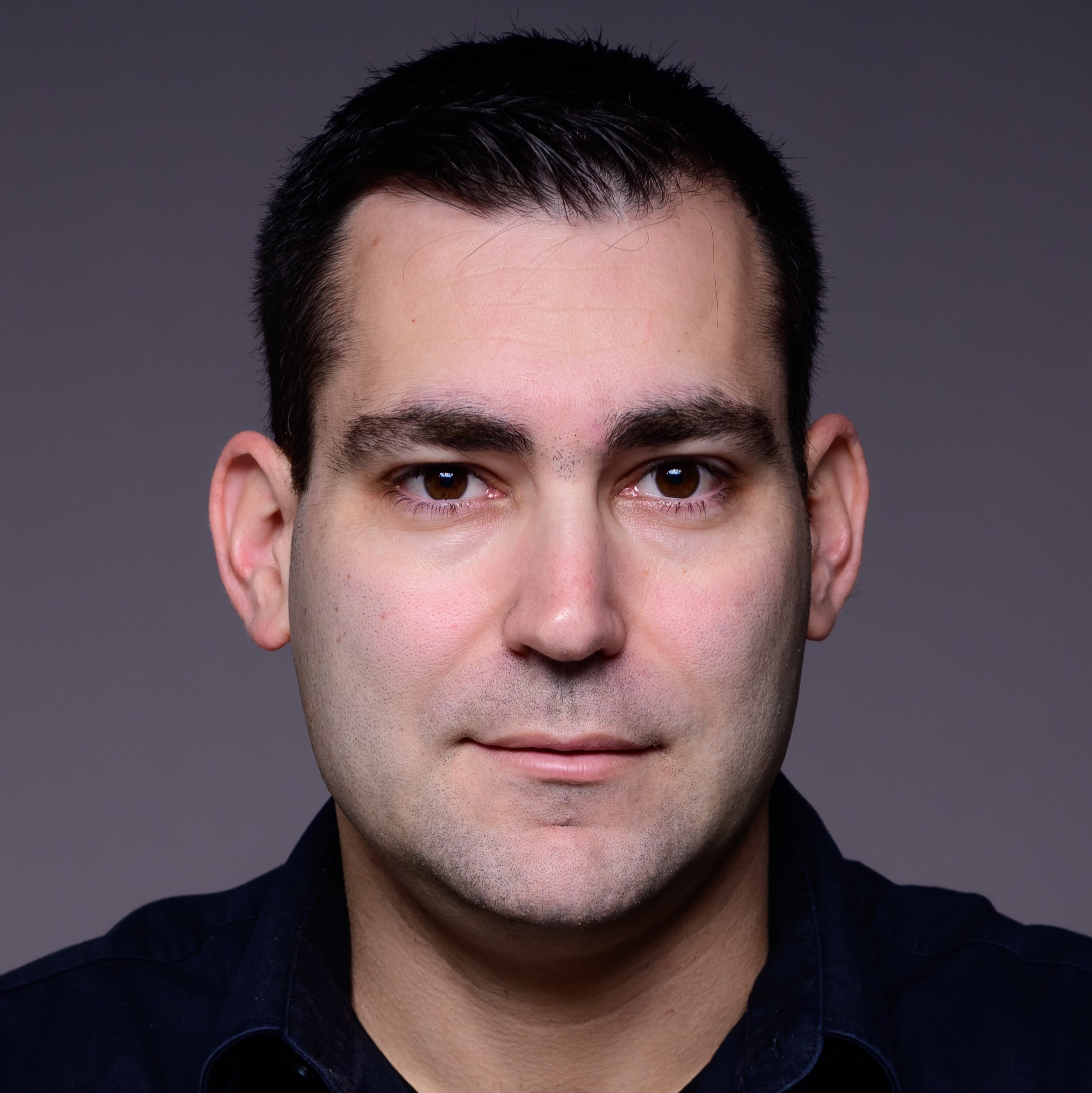}}]{Barak Or} (Member, IEEE) received a B.Sc. degree in aerospace engineering (2016), a B.A. degree (cum laude) in economics and management (2016), and an M.Sc. degree in aerospace engineering (2018) from the Technion–Israel Institute of Technology. He graduated with a Ph.D. degree from the University of Haifa, Haifa (2022).
He founded ALMA Technologies Ltd (2021), focusing on Navigation and machine learning algorithms. 
His research interests include navigation, deep learning, sensor fusion, and estimation theory.
\end{IEEEbiography}

\end{document}